\documentclass[letterpaper, 10 pt, conference]{ieeeconf}  

\IEEEoverridecommandlockouts                              

\usepackage{amsfonts}
\usepackage{amsmath}
\usepackage{algorithm}
\usepackage{algorithmicx}

\usepackage[noend]{algpseudocode}
\usepackage{mathtools}
\usepackage{array}
\usepackage{graphics}
\usepackage{wrapfig}
\usepackage{subcaption}
\usepackage{float} 
\usepackage{siunitx} 

\usepackage{cleveref}
\usepackage{epsfig}
\usepackage{todonotes}
\usepackage{lipsum}
\usepackage{cite}
\usepackage{todonotes}

\DeclareMathOperator*{\argmax}{arg\,max}

\DeclareMathAlphabet\mathbfcal{OMS}{cmsy}{b}{n}

\title{\LARGE \bf
Learning Multi-Robot Decentralized Macro-Action-Based Policies \\ 
via a Centralized Q-Net
}

\author{Yuchen Xiao, Joshua Hoffman, Tian Xia and Christopher Amato
\thanks{Khoury College of Computer Sciences, Northeastern University, Boston, MA 02115, USA \{xiao.yuch, hoffman.j, xia.tia\}@husky.neu.edu, c.amato@northeastern.edu}
}

\begin{document}

\maketitle
\thispagestyle{empty}
\pagestyle{empty}

\begin{abstract}
    In many real-world multi-robot tasks, high-quality solutions often require a team of robots to perform asynchronous actions under decentralized control. Decentralized multi-agent reinforcement learning methods have difficulty learning decentralized policies because of the environment appearing to be non-stationary due to other agents also learning at the same time. In this paper, we address this challenge by proposing a macro-action-based decentralized multi-agent double deep recurrent Q-net (MacDec-MADDRQN) which  trains each decentralized Q-net using a centralized Q-net for action selection. A generalized version of MacDec-MADDRQN with two separate training environments, called Parallel-MacDec-MADDRQN, is also presented to leverage either centralized or decentralized exploration. The advantages and the practical nature of our methods are demonstrated by achieving near-centralized results in simulation and having real robots accomplish a warehouse tool delivery task in an efficient way.       

\end{abstract}

\section{INTRODUCTION}

\begin{figure}[t!]
    \centering
    \subcaptionbox{\label{intro_a}}
        [0.46\linewidth]{\includegraphics[width=4.2cm, height=2.2cm]{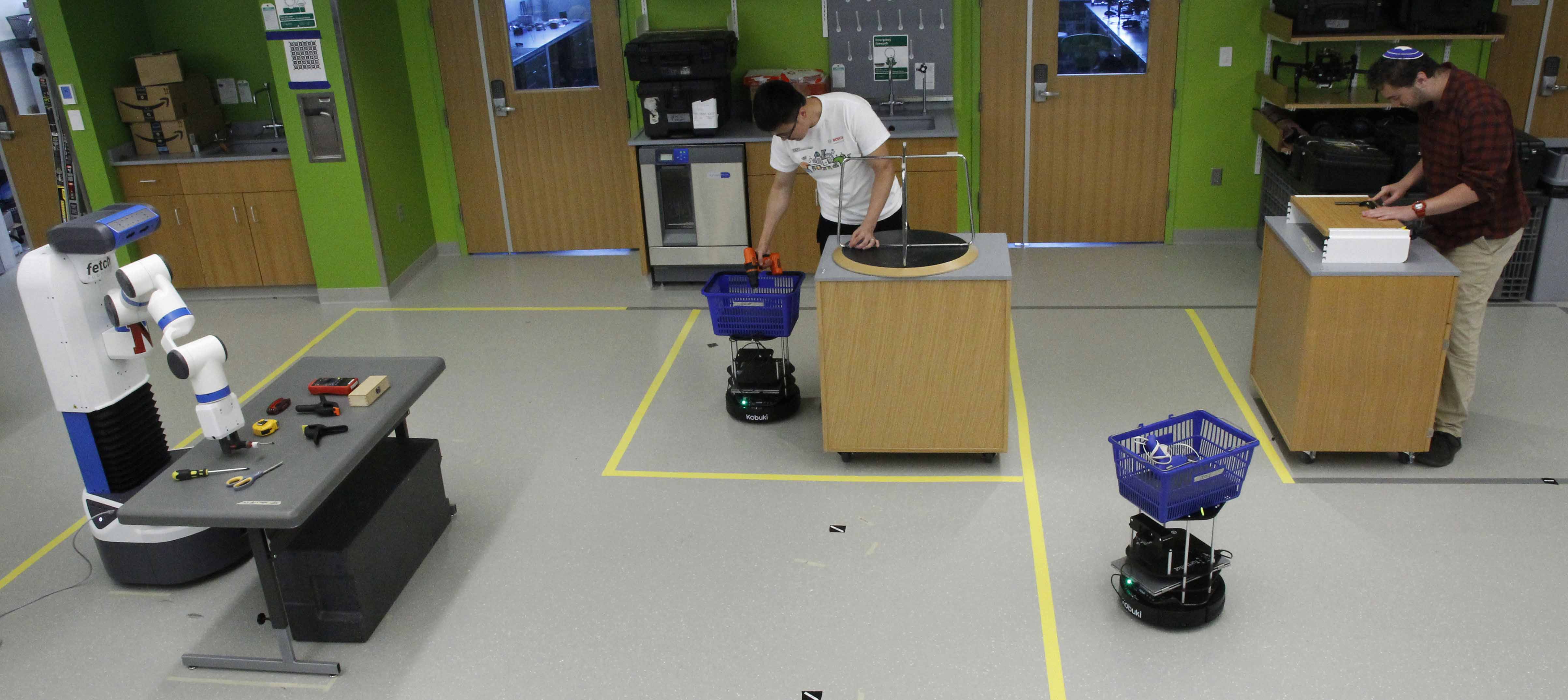}}
    ~
    \centering
    \subcaptionbox{\label{intro_b}}
        [0.22\linewidth]{\includegraphics[width=2.1cm, height=2.2cm]{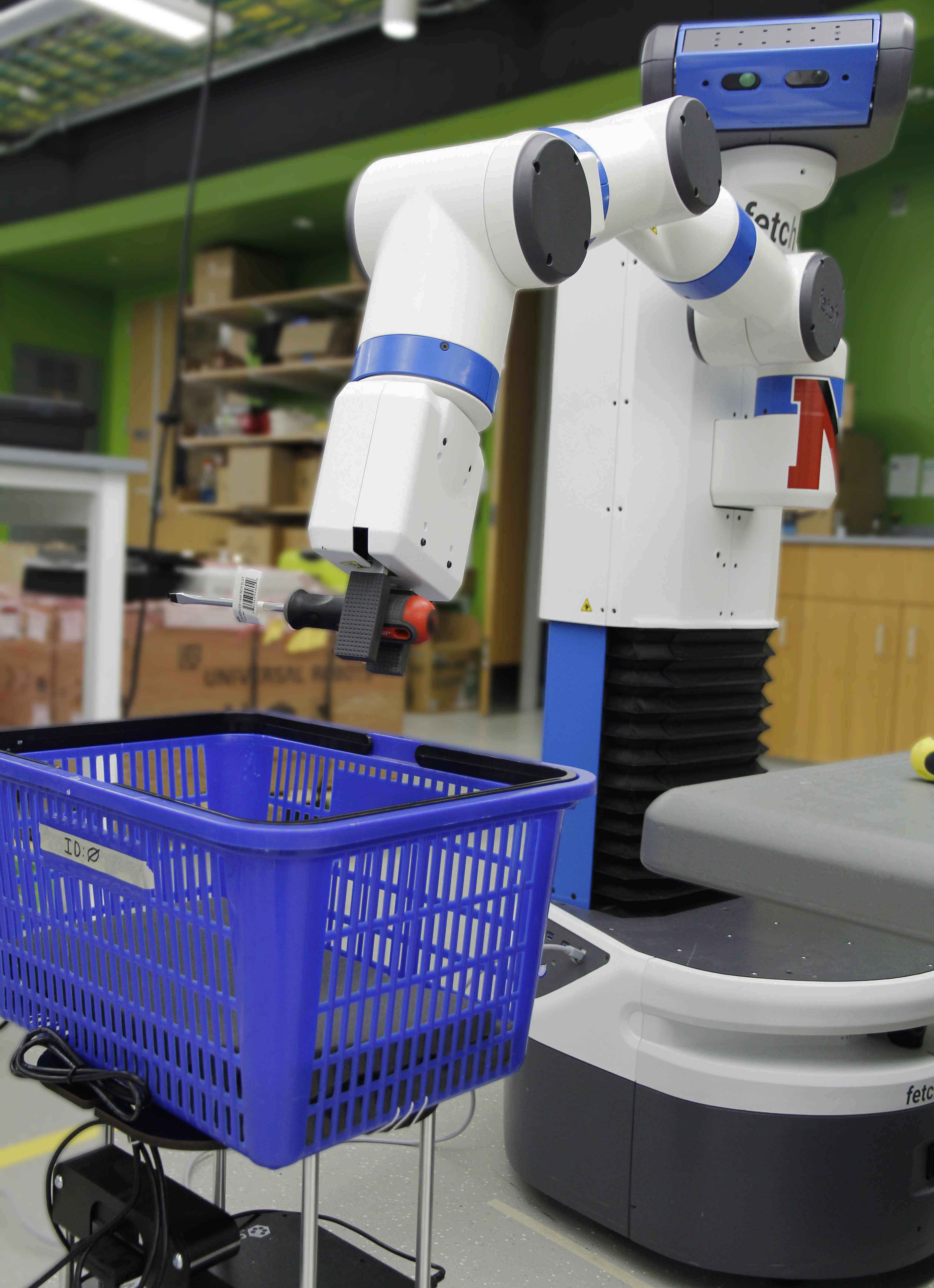}}
    ~
    \centering
    \subcaptionbox{\label{intro_c}}
        [0.21\linewidth]{\includegraphics[width=2.0cm, height=2.2cm]{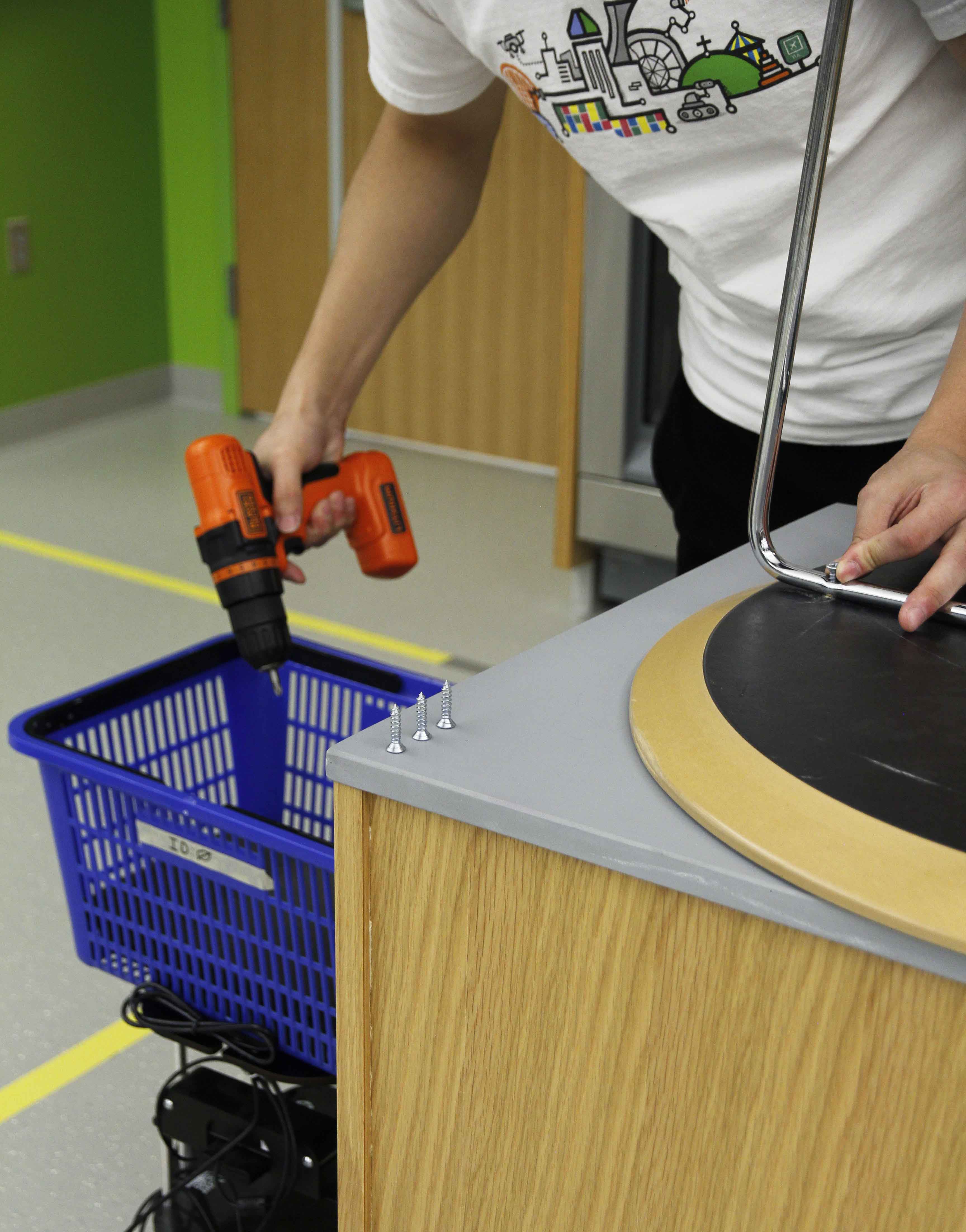}}
    \vspace{-2mm}
    \caption{Warehouse tool delivery task: (a) Three robots deliver tools to two humans; (b) Collaborative tool passing; (c) Correct tool delivered.} 
    \vspace{-6mm}
    \label{intro}
\end{figure}

Multi-robot systems have become ubiquitous in our daily lives, such as drones for applications such as agricultural inspection, warehouse robots, and self-driving cars~\cite{agricultural,kiva,Waymo}. 
For example, consider a warehouse environment (Fig.~\ref{intro_a}), where a Fetch robot~\cite{Wise:M} and two Turtlebots~\cite{Turtlebot} are autonomously delivering tools in order to assist two humans with their assembly tasks. To be more efficient, the robots should be able to predict which tool the human workers will potentially need rather than always waiting for a human's request, while collaborating with the other robots to find the tool in advance and passing it to one of the Turtlebots (Fig.~\ref{intro_b}) for delivery (Fig.~\ref{intro_c}). Performing these high-quality coordination behaviors in large, stochastic and uncertain environments is challenging for the robots, because it requires the robots to operate asynchronously according to local information while reasoning about cooperation between teammates.   

Although, several multi-agent deep reinforcement learning approaches have been proposed and have achieved high-quality performance \cite{DecHDRQN,foerster:aaai18,lowe2017multi,rashid:icml18,Sunehag}, these methods assume synchronized primitive actions. 
Our very recent work~\cite{YuchenCoRL} bridged this gap by proposing the first asynchronous macro-action-based multi-agent deep reinforcement learning frameworks. Macro-actions naturally represent temporally extended robot controllers that can be executed in an asynchronous manner~\cite{AAMAS14AKK,AmatoJAIR19}. In that paper, we proposed approaches for both learning decentralized macro-action-value functions and centralized joint-macro-action-value functions. However, the decentralized method, using Decentralized Hysteretic DRQN with Double DQN (Dec-HDDRQN), performed poorly in large and complex domains. Nevertheless, decentralized execution is necessary for cases when there is limited or no communication between robots. 


In this paper, we improve the learning of decentralized policies via two contributions: (a) A new macro-action-based decentralized multi-agent deep double-Q learning approach, called MacDec-MADDRQN, which adopts \emph{centralized training with decentralized execution} by allowing each individual decentralized Q-net update to use a centralized Q-net; (b) MacDec-MADDRQN introduces a choice of $\epsilon$-greedy exploration, either based on the centralized Q-net or the decentralized Q-nets. The best choice is often not clear without knowledge of domain properties. Therefore, a more general version, called Parallel-MacDec-MADDRQN, is proposed, in which, the centralized Q-net is trained purely based on the experiences generated by using centralized $\epsilon$-greedy exploration in one environment, simultaneously, agents perform decentralized exploration in a separate environment, and each decentralized Q-net is then optimized using the decentralized data and the centralized Q-net. 

We evaluate our methods in both simulation and in hardware. In simulation, our methods outperform the previous decentralized method by either converging to a much higher value or learning faster in both a benchmark domain and a Warehouse Tool Delivery domain with a single human involved. We also deploy the decentralized policies learned in simulation on real robots which shows high-quality cooperation to deliver the correct tools in an efficient way. To our knowledge, 
this is the first instance of running a set of decentralized macro-action-based policies that were trained via deep reinforcement learning on a team of real robots.









\section{BACKGROUND}

We first discuss macro-action-based Dec-POMDPs~\cite{AAMAS14AKK,AmatoJAIR19} and deep Q-learning, and then provide an overview of our previous related approach~\cite{YuchenCoRL}. 

\subsection{MacDec-POMDPs}

Decentralized fully cooperative multi-agent decision-making under uncertainty can be modeled as a decentralized POMDP (Dec-POMDP)~\cite{Oliehoek}. Due to the assumption of synchronous actions that require the same amount of time for each agent, Dec-POMDPs are not applicable to multi-robot planning and learning scenarios in real-world. MacDec-POMDPs, formalized by introducing macro-actions into Dec-POMDPs, inherently allow asynchronous execution among robots with temporally extended macro-actions that can begin and end at different times for each agent. 

Formally, a MacDec-POMDP is defined as a tuple $\langle I, S, A, \Omega, M, \zeta, O, T, Z, R\rangle$, where $I$ is a finite set of agents; $S$ is a finite set of environment states; $A \equiv \times_iA_i$ and $\Omega\equiv\times_i\Omega_i$ are the spaces of joint-primitive-action and joint-primitive-observation respectively; $M\equiv\times_iM_i$ is the joint set of each agent's finite macro-action space $M_i$; $\zeta\equiv\times_i\zeta_i$ is the set of joint macro-observations over agents' finite macro-observation space $\zeta_i$. Given a macro-action-based policy, each agent $i$ is allowed to asynchronously choose a macro-action $m_i=\langle\beta^{m}, I^{m}, \pi^{m} \rangle_i$ that depends on individual macro-action-observation histories, where $\beta^{m}:H^A_i\rightarrow[0,1]$ is the stochastic termination condition and $I^{m}\subset H^{M}_i$ is the initiation set of the corresponding macro-action $m_i$, respectively depending on the primitive-action-observation history space $H^A_i$ and macro-action-observation history space $H^M_i$ of agent $i$; $\pi^{m}:H^{A}_i\rightarrow A_i$ denotes the low-level policy to achieve the macro-action $m$, and during the execution, each agent's primitive-observation $o_i\in\Omega_i$ is generated according to probability observation function $O_i(o_i, a_i, s)=\Pr(o_i\mid a_i,s)$, and a shared immediate reward $r(s,\vec{a})$, where $\vec{a}\in A \equiv \times_iA_i$, is issued according to the reward function $R:S\times A\rightarrow\mathbb{R}$. Importantly, considering the stochastic terminations and the asynchronous executions of macro-actions over agents, the transition function is defined as $T(s',\vec{\tau}, s, \vec{m}) = \Pr(s',\vec{\tau}\mid s, \vec{m})$, where $\vec{\tau}$ is the time-step at which \emph{any agent i} completes its macro-action $m_i$, and also indicates the termination of the joint-macro-action $\vec{m}$; 
Successively, a new joint-macro-observation $\vec{z}\in\zeta\equiv\times_i\zeta_i$ is generated based on the macro-observation function $Z(\vec{z}, \vec{m},s') = \Pr(\vec{z}\mid \vec{m},s')$; 
Note that, each agent keeps updating its primitive observation every time-step, but only updates macro-observation when its current macro-action has terminated. The objective is to optimize the joint high-level policy $\Psi=\times_i\Psi_i$ such that the following expected discounted return from an initial state $s_0$ is maximized:

\begin{equation}
    \Psi^*=\argmax_{\Psi}\mathbb{E}\Biggl[\sum_{t=0}^{h-1}\gamma^tr(s_{(t)},\vec{a}_{(t)})\mid s_{(0)}, \pi, \Psi\Biggr]
\end{equation}


\subsection{Deep Recurrent Q-Network and Double DQN}

Deep Q-learning is a state-of-the-art approach using a deep Q-network (DQN) parameterized by $\theta$ as an action-value approximator which is iteratively updated to minimize the loss: $\mathcal{L}(\theta)=\mathbb{E}_{<s, a, r, s'>\sim\mathcal{D}}\Big[\big(y - Q_{\theta}(s, a)\big)^2 \Big]$, where $ y=r + \gamma\argmax_{a'}Q_{\theta^-}(s',a')$. Experience replay and a less frequently updated target Q-network, parameterized by $\theta^-$, are employed for improving performance and stabilizing learning\cite{DQN}. DQN with a recurrent layer (DRQN) has been widely adopted in partial observable domains to allow agent's actions to depend on abstractions of action-observation histories rather than states (or a single observation)~\cite{DRQN}. Double DQN incorporates double Q-learning \cite{DoubleQ} into DQN to provide an unbiased target action-value estimation, $ y=r + \gamma Q_{\theta^-}(s', \argmax_{a'}Q_{\theta}(s',a'))$ \cite{DDQN}, which leverages the above two Q-networks. In this paper, we mainly compare our decentralized learning approaches with Decentralized Hysteretic DRQN (which uses two learning rates for more robust updating against negative TD error) with Double DQN (Dec-HDDRQN)~\cite{DecHDRQN}, and centralized learning via Double DRQN (Cen-DDRQN).

\subsection{Learning Macro-Action-Based Deep Q-Nets}
\label{LMADQN}

Although there has been several popular multi-agent deep reinforcement learning methods achieving impressive performance in cooperative as well as competitive domains~\cite{DecHDRQN, Sunehag, rashid:icml18, foerster:aaai18, lowe2017multi}, they all require primitive actions and synchronous action execution. There was no principled way to utilize these methods to learn macro-action-based policies, in which the challenges were how to properly update macro-action values and correctly maintain macro-action-observation trajectories.   

To cope with the above challenges, in our previous work~\cite{YuchenCoRL}, we first proposed a decentralized macro-action-based learning method that is based on Dec-HDDRQN with a new buffer called Macro-Action Concurrent Experience Reply Trajectories (Mac-CERTs). This buffer contains the macro-action-observation experience represented as a tuple $\langle z_i, m_i, z'_i, r^c_i\rangle$ for each agent $i$, where $r^c_i= \sum_{t=t_{m_i}}^{\tau} r_t$ is an accumulated reward for the macro-action $m_i$ from its beginning time-step $t_{m_i}$ to the termination step $\tau$. In the training phase, each agent individually updates its own macro-action-value function $Q_{\theta_i}(h_i,m_i)$, using a concurrent mini-batch of sequential experiences sampled from Mac-CERTs, by minimizing the loss:  $\mathcal{L}(\theta_i)=\mathbb{E}_{<z_i, m_i, z'_i, r^c_i>\sim\mathcal{D}}\Big[\bigl(y_i - Q_{\theta _i}(h_i, m_i)\bigr)^2 \Big]$, where $y_i = r^c_i + \gamma Q_{\theta_i^-}\bigl(h'_i, \argmax_{m'_i} Q_{\theta_i}(h'_i,m'_i)\bigr)$ and $h_i$ denotes the macro-action-observation history of agent $i$.  
For cases when a centralized macro-action-based policy is possible, we also proposed a novel centralized replay buffer called Macro-Action Joint Experience Replay Trajectories (Mac-JERTs)~\cite{YuchenCoRL}. At each execution step, this buffer collects a joint macro-action-observation experience represented as a tuple $\langle \vec{z}, \vec{m}, \vec{z}\,', \vec{r}\,^c\rangle$, where $\vec{r}\,^c=\sum_{t=t_{\vec{m}}}^{\vec{\tau}} r_t$ is a shared joint accumulated reward for the agents' joint macro-action $\vec{m}$ from its beginning time-step $t_{\vec{m}}$ to the ending time-step $\vec{\tau}$ when \emph{any agent} terminates its macro-action. The centralized macro-action-value function $Q_\phi(\vec{h},\vec{m})$ is then optimized by minimizing the loss: $\mathcal{L}(\phi)=\mathbb{E}_{<\vec{z}, \vec{m}, \vec{z}\,', \vec{r}\,^c>\sim\mathcal{D}}\Big[\bigl(y - Q_{\phi}(\vec{h}, \vec{m})\bigr)^2 \Big]$, where $y = \vec{r}\,^c + \gamma Q_{\phi-}\bigl(\vec{h}\,', \argmax_{\vec{m}\,'} Q_{\phi}(\vec{h}\,', \vec{m}\,'\mid \vec{m}^{\text{undone}})\bigr)$. Here, $\vec{m}^{\text{undone}}$ is the joint-macro-action over the agents who have not completed their macro-actions in the sampled experience. 
Note that, this conditional operation considers the agents' asynchronous macro-action execution status which is accessible from Mac-JERTs during training.   

Building on our previous work, in this paper, we extend Double DQN to decentralized multi-agent macro-action-based policy learning under partial observability in the manner of \emph{centralized training with decentralized execution}. In this new method, each agent is able to update its own Q-net by taking into account the effects of other agents' behaviors in the environment, naturally surmounting the non-stationary environment issue from each agent's perspective.   

\section{APPROACH}

In multi-agent environments, decentralized learning causes non-stationarity from each agent's perspective as other agents policies change during learning. Learning a centralized joint-action-value function to guide each agent's decentralized policy updating has been being a very popular training manner to conquer the non-stationarity~\cite{foerster:aaai18, lowe2017multi}. VDN and QMIX also use centralized training by first training a centralized, but factored, Q-net that is decomposed into a decentralized Q-net for each agent for use in execution~\cite{Sunehag,rashid:icml18}. In this section, we propose a new multi-agent Double DQN-based approach, called MacDec-MADDRQN, to learn decentralized macro-action-value functions that are trained with a centralized joint macro-action-value function.  

\subsection{Macro-Action-Based Decentralized Multi-Agent Double Deep Recurrent Q-Net (MacDec-MADDRQN)}

Double DQN has been implemented in multi-agent domains for learning either centralized or decentralized policies~\cite{MADDQN,WDDQN, YuchenCoRL}. However, in the decentralized learning case, each agent independently adopts double Q-learning purely based on its own local information. 
Learning only from local information often impedes agents from achieving high-quality cooperation.    

In order to take advantage of centralized information for learning decentralized Q-networks, we train the centralized joint macro-action-value function $Q_{\phi}$ and each agent's decentralized macro-action-value function $Q_{\theta_i}$ simultaneously, and the target value for updating decentralized macro-action-value function $Q_{\theta_i}$ is then calculated by using the centralized $Q_{\phi}$ for macro-action selection and the decentralized target-net $Q_{\theta_i^-}$ for value estimation. 

More concretely, consider a domain with $N$ agents, and both the centralized Q-network $Q_{\phi}$ and decentralized Q-networks $Q_{\theta_i}$ for each agent $i$ are represented as DRQNs~\cite{DRQN}. The experience replay buffer $\mathbfcal{D}$, a merged version of Mac-CERTs and Mac-JERTs, contains the tuples $\langle\mathbf{z}, \mathbf{m}, \mathbf{z'}, \mathbf{r^c}, \vec{r}^{\,c}\rangle$, where $\mathbf{z} = \{z_0,..., z_N\}$, $\mathbf{m}=\{m_0,...,m_N\}$ and $\mathbf{r^c}=\{r^c_0,...,r^c_N\}$. In each training iteration, agents sample a mini-batch of sequential experiences to first optimize the centralized joint macro-action-value function $Q_\phi$ in the way mentioned in Section~\ref{LMADQN}, and then update each decentralized macro-action-value function by minimizing the squared TD error:        
\vspace{-1mm}
\begin{equation}
    \mathcal{L}(\theta_i)=\mathbb{E}_{<\mathbf{z}, \mathbf{m}, \mathbf{z'}, \mathbf{r^c}, \vec{r}^{\,c} >\sim\mathbfcal{D}}\Big[\big(y_i - Q_{\theta_i}(h_i, m_i)\big)^2 \Big] 
\end{equation}
where,
\vspace{-2mm}
\begin{equation}
    y_i=r^c_i + \gamma Q_{\theta^-_i}\biggr[h_i',\big[\argmax_{\mathbf{m'}}Q_{\phi}(\mathbf{h'}, \mathbf{m'})\big]_i\biggr]
    \label{unCondi}
\end{equation}

In Eq.~\ref{unCondi}, $\big[\argmax_{\mathbf{m'}}Q_{\phi}(\mathbf{h'}, \mathbf{m'})\big]_i$ implies selecting the joint macro-action with the highest value and then selecting the individual macro-action for agent $i$. In this updating rule, not only are double estimators $Q_{\theta_i^-}$ and $Q_{\phi}$  applied to counteract overestimation on target Q-values, but also a centralized heuristic on action selection is embedded. Now, from each agent's perspective, the target Q-value is calculated by assuming all agents will behave based on the centralized Q-net next step (Eq.~\ref{unCondi}), in which the provided global information by the centralized Q-net will help each agent to avoid getting trapped in local optima and also facilitates them to learn cooperation behaviors.      

Additionally, similar to the idea of the conditional operation for training a centralized joint macro-action-value function discussed in Section~\ref{LMADQN}, in order to obtain a more accurate prediction by taking each agent's macro-action executing status into account, Eq.~\ref{unCondi} can be rewritten as:

\vspace{-3mm}
\begin{equation}
    y_i=r^c_i + \gamma Q_{\theta^-_i}\biggr[h_i',\big[\argmax_{\mathbf{m'}}Q_{\phi}(\mathbf{h'}, \mathbf{m'}\mid \mathbf{m^{undone}})\big]_i\biggr]
    \label{Condi}
\end{equation}

\subsection{$\epsilon$-greedy Exploration Policy Selection}

Exploration is also a difficult problem in multi-agent reinforcement learning. 
$\epsilon$-greedy exploration has been widely used in many methods such as Q-learning to generate training data~\cite{Sutton1998}. In DQN-based methods, as a hyper-parameter, $\epsilon$ often acts with a linear decay along with the training steps from $1.0$ to a lower value to achieve the trade-off between exploration and exploitation. And, exploration can be done based on either the centralized or decentralized policies. Centralized exploration may help to choose cooperative actions more often that would have a low probability of being selected from decentralized policies, and decentralized exploration may provide more realistic data that is actually achievable by decentralized policies.      

Therefore, in our approach, besides tuning $\epsilon$, we introduce a hyper-selection for performing a $\epsilon$-greedy behavior policy that can perform either centralized exploration based on $Q_\phi$ or decentralized exploration using each agent's $Q_{\theta _i}$.  

\begin{algorithm}[t]
    \footnotesize
    \caption{Parallel-MacDec-MADDRQN}
    \label{alg1}
        \begin{algorithmic}
            \State Initialize centralized Q-Networks: $Q_\phi$, $Q_\phi^-$
            \State Initialize decentralized Q-Networks for each agent $i$: $Q_{\theta_i}$, $Q_{\theta_i}^-$
            \State Initialize two parallel environments \emph{cen-env}, \emph{dec-env}
            \State Initialize two step counters $t_{\text{cen-env}}$, $t_{\text{dec-env}}$
            \State Initialize centralized buffer $\mathcal{D}_{\text{cen}} \leftarrow$ Mac-JERTs
            \State Initialize decentralized buffer $\mathcal{D}_{\text{dec}} \leftarrow$ Mac-CERTs
            \State Get initial joint-macro-observation $\vec{z}$ for agents in \emph{cen-env}
            \State Get initial macro-observation $z_i$ for each agent $i$ in \emph{dec-env}
            \For{\emph{dec-env-episode} = $1$ to $M$}
                \State Agents take joint-macro-action with cen-$\epsilon$-greedy using $Q_{\phi}$
                \State Store $\langle \vec{z}, \vec{m}, \vec{z}{\,}',\vec{r}^{\,c}\rangle$ in $\mathcal{D}_{\text{cen}}$
                \State $t_{\text{cen-env}}\leftarrow t_{\text{cen-env}} + 1$
                \State Each agent $i$ takes macro-action with dec-$\epsilon$-greedy using $Q_{\theta_i}$
                \State Store $\langle z_i, m_i, z_i',r^{c}_i\rangle$ in $\mathcal{D}_{\text{dec}}$
                \State $t_{\text{dec-env}}\leftarrow t_{\text{dec-env}} + 1$
                \If{$t_{\text{dec-env}}$ mod $I_{\text{train}} == 0$}
                    \State Sample a mini-batch $\mathcal{B}_{\text{cen}}$ of sequential experiences 
                    \State $\langle \vec{z}, \vec{m}, \vec{z}{\,}',\vec{r}^{\,c}\rangle$ from $\mathcal{D}_{\text{cen}}$
                    \State Perform a gradient decent step on $\bigr(y-Q_\phi(\vec{h}, \vec{m})\bigr)^2_{\mathcal{B}_{\text{cen}}}$, where
                    \State $y = \vec{r}\,^c + \gamma Q_{\phi-}\bigl(\vec{h}\,', \argmax_{\vec{m}\,'} Q_{\phi}(\vec{h}\,', \vec{m}\,'\mid \vec{m}^{\text{undone}})\bigr)$.
                    \State Sample a mini-batch $\mathcal{B}_{\text{dec}}$ of sequential experiences 
                    \State $\langle z_i, m_i, z_i',r^{\,c}_i\rangle$ from $\mathcal{D}_{\text{dec}}$ for each agent $i$ 
                    \State Perform a gradient decent step on $\bigr(y_i-Q_{\theta_i}(h_i, m_i)\bigr)^2_{\mathcal{B}_{\text{dec}}}$, where
                    \State $y_i=r^c_i + \gamma Q_{\theta^-_i}\biggr[h_i',\big[\argmax_{\mathbf{m'}}Q_{\phi}(\mathbf{h'}, \mathbf{m'}\mid \mathbf{m^{undone}})\big]_i\biggr]$

                \EndIf
                \State \textbf{end if}
                \If{$t_{\text{dec-env}}$ mod $I_{\text{TargetUpdate}} = 0$}
                    \State Update centralized target network $\phi^-\leftarrow\phi$
                    \State Update each agent's decentralized target network $\theta_i^-\leftarrow\theta_i$
                \EndIf
                \State \textbf{end if}
                \If{$t_{\text{cen-env}}= \,$max-episode-length \textbf{or} terminal state}
                    \State Reset \emph{cen-env} 
                    \State Get initial joint-macro-observation $\vec{z}$ for agents in \emph{cen-env}
                \EndIf
                \State \textbf{end if}
                \If{$t_{\text{dec-env}}= \,$max-episode-length \textbf{or} terminal state}
                    \State Reset \emph{dec-env} 
                    \State Get initial macro-observation $z_i$ for each agent $i$ in \emph{dec-env}
                \EndIf
                \State \textbf{end if}
            \EndFor
            \State \textbf{end for}

        \end{algorithmic}
\end{algorithm}

However, without having enough knowledge about the properties of a given domain in the very beginning, it is not clear which exploration choice is the best. To cope with this, we propose a more generalized version of MacDec-DDRQN, called \emph{Parallel-MacDec-MADDRQN}, summarized in Algorithm~\ref{alg1}. The core idea is to have two parallel environments with agents respectively performing centralized exploration (cen-$\epsilon$-greedy) and decentralized exploration (dec-$\epsilon$-greedy) in each. The centralized $Q_\phi$ is first trained purely using the centralized experiences, while each agent's decentralized $Q_{\theta_i}$ is then optimized using Eq.~\ref{Condi} with only decentralized experiences. The performance of this algorithm in the Warehouse domain is presented in Section~\ref{SimResults} 

\section{SIMULATION EXPERIMENTS}

In this section, we describe two macro-action-based multi-robot domains designed in our previous work~\cite{YuchenCoRL}, the Box Pushing (BP) domain and the Warehouse Tool Delivery (WTD) domain. We evaluate our approaches in these two domains while comparing with macro-action-based Dec-HDDRQN, fully centralized training via DDRQN (Cen-DDRQN), and some ablations we consider. 

\subsection{Domain Setup}
\label{Domains}

\begin{figure}[h!]
    \vspace{-2mm}
    \centering
    \begin{subfigure}{.2\textwidth}
        \centering
        \includegraphics[height=2.3cm]{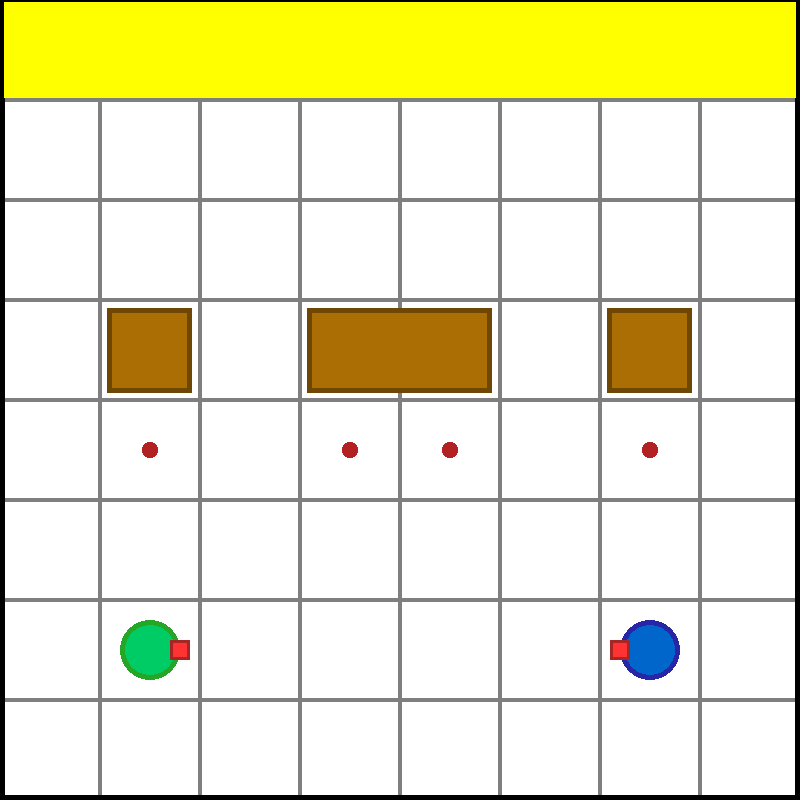}
        \caption{Box Pushing}
        \label{bpma}
    \end{subfigure}
    \begin{subfigure}{.25\textwidth}
        \centering
        \includegraphics[height=2.3cm]{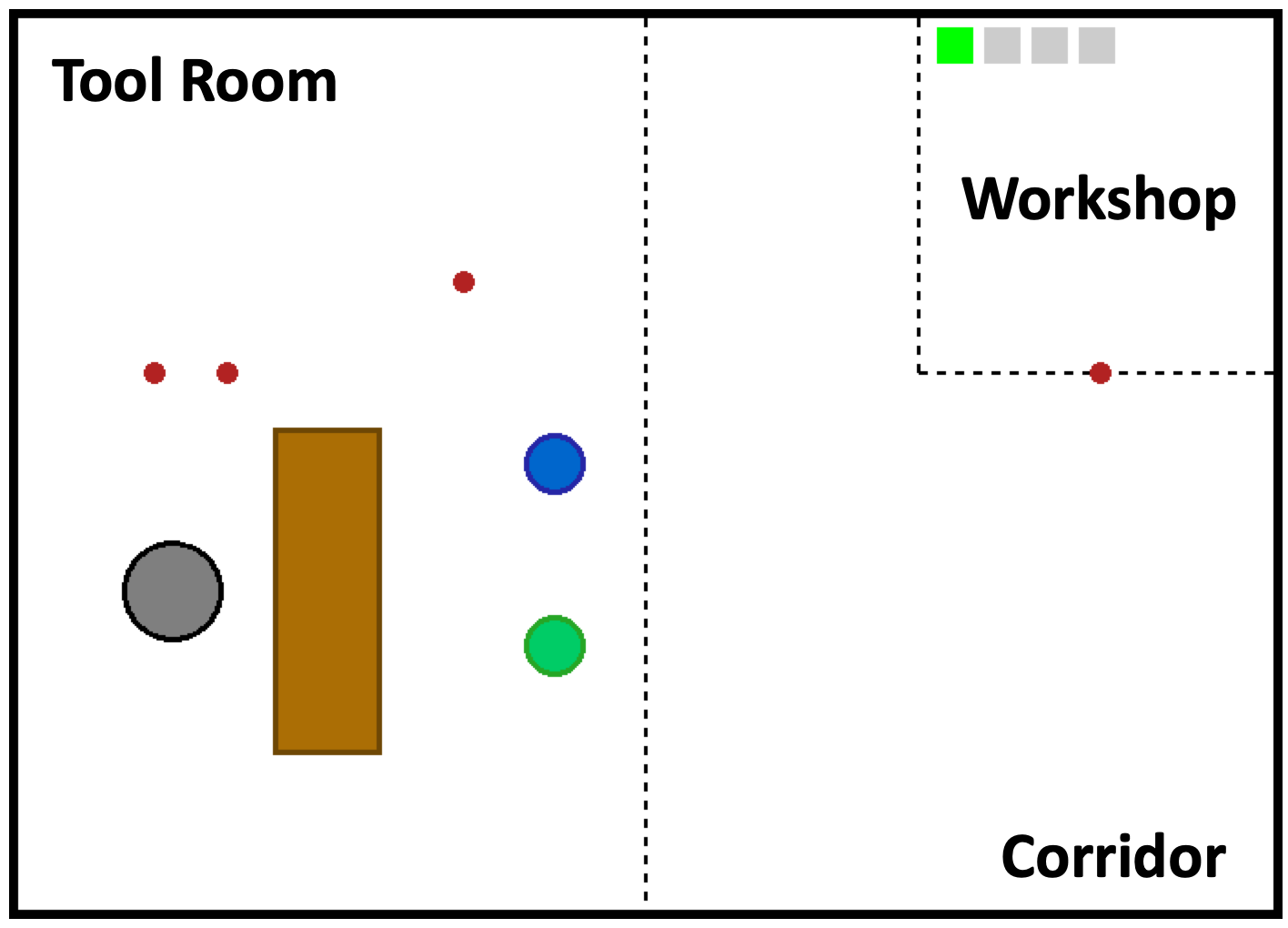}
        \caption{Warehouse Tool Delivery}
        \label{WTD}
    \end{subfigure}
    \vspace{-1mm}
    \caption{Experimental environments in simulation}
    \vspace{-2mm}
    \label{domains}
\end{figure}

        \subsubsection{Box Pushing} (Fig.~\ref{bpma}). This domain has two mobile robots with the goal of cooperatively pushing a big box (middle brown square), which is only movable when two robots push it together, to the goal area (yellow bar at the top). The difficulties come from partial observability (each robot is only allowed to observe one cell in front) and two small boxes which attract the robots to learn the sub-optimal policy that is pushing one small box on its own. We provide two categories of macro-actions for each robot: (a) One step macro-actions, {\bf\emph{Turn-left, Turn-right}} and {\bf \emph{Stay}}; (b) Long-term macro-actions, {\bf\emph{Move-to-small-box(i)}} and {\bf\emph{Move-to-big-box(i)}}, navigate the robot to one of the red waypoints below the corresponding box and ends with the robot facing it; {\bf\emph{Push}} commands the robot to keep moving forward until arriving at the boundary of the grid world, touching the big box, or pushing a small box to the goal area. 
The macro-observation space for each robot consists of four different possible values associated with the cell in front of the robot: \emph{empty}, \emph{teammate}, \emph{boundary}, \emph{small box} and \emph{big box}. Robots obtain $+100$ or $+10$ rewards respectively for pushing the big box or a small box to the goal area, and a $-5$ penalty is assigned to the team when any robot pushes the big box alone or hits the boundaries. Robots also get $-0.1$ reward per time-step. Note that each episode terminates either by reaching the horizon limitation or when any box pushed to the goal area.  

\subsubsection{Warehouse Tool Delivery} (Fig.~\ref{WTD}). In order to test if our approach is applicable to address real-world industrial problems, we developed  a tool delivery task for a warehouse environment ($5\times7$ continuous space), in which, one human works on an assembly task (4 steps in total and each step takes 18 units time) in the workshop. The human always starts from step one and needs a particular tool for each future step to continue. The objective of the three robots is to assist the human to finish the assembly task as soon as possible by collaboratively searching for the right tools in the proper order on the brown table 
 and then passing them to one of the mobile robots (the green or blue) to accomplish the delivery in time. To make this problem more challenging, the info about the correct tool that the human needs for future step is not known to the robots, so it has to be learned during training. Also, the human is only allowed to possess one tool at a time from the mobile robots. 

Each mobile robot has three macro-actions: {\bf\emph{Go-to-WS}} navigates the robot to the red waypoint at the workshop; {\bf\emph{Go-to-TR}} drives the robot to the upper right waypoint in the tool room; the duration of these two macro-actions depends on the robot's moving speed (0.6 in our case); {\bf\emph{Get-Tool}} navigates the robot to the pre-assigned waypoint beside the table and waits there until either obtaining one tool from the gray robot or 10 time-steps have passed. Also, there are four applicable macro-actions for the gray robot: {\bf\emph{Wait-M}} lasts 1 time-step; {\bf\emph{Search-Tool(i)}} takes 6 time-steps to find the tool $i$ and place it in the staging area 
(lower left on the table where can hold at most  two tools). 
Running this action when the staging area is fully occupied leads the robot to pause for 6 time-steps. {\bf\emph{Pass-to-M(i)}} lasts 4 time-steps to pass one of the tools from the staging area, in first-in-first-out order, to mobile robot $i$.

We allow each mobile robot to capture four different features in a macro-observation, including \emph{location}, \emph{the human's current step} (only accessible when in the workshop), \emph{the tools being carried} by that robot, and \emph{the number of tools} in the staging area (only observable when in the tool room). While, the gray robot can monitor \emph{which mobile robot} is beside the table and \emph{the number of tools} in the staging area. 

The global rewards provide $-1$ each time-step to encourage the robots to deliver the object(s) in a timely manner without causing the human to pause; a penalty of $-10$ is given when the gray robot executes {\bf\emph{Pass-to-M(i)}} but no mobile robots are beside the table; a bonus of $+100$ is awarded to the entire team when the robots successfully deliver a correct tool to the human.

\subsection{Results in the Box Pushing Domain}
\label{SimResults}

\begin{figure}[t!]
    \centering
    \subcaptionbox{Grid world $10\times10$\label{bpma10}}
        [0.47\linewidth]{\includegraphics[height=3.1cm]{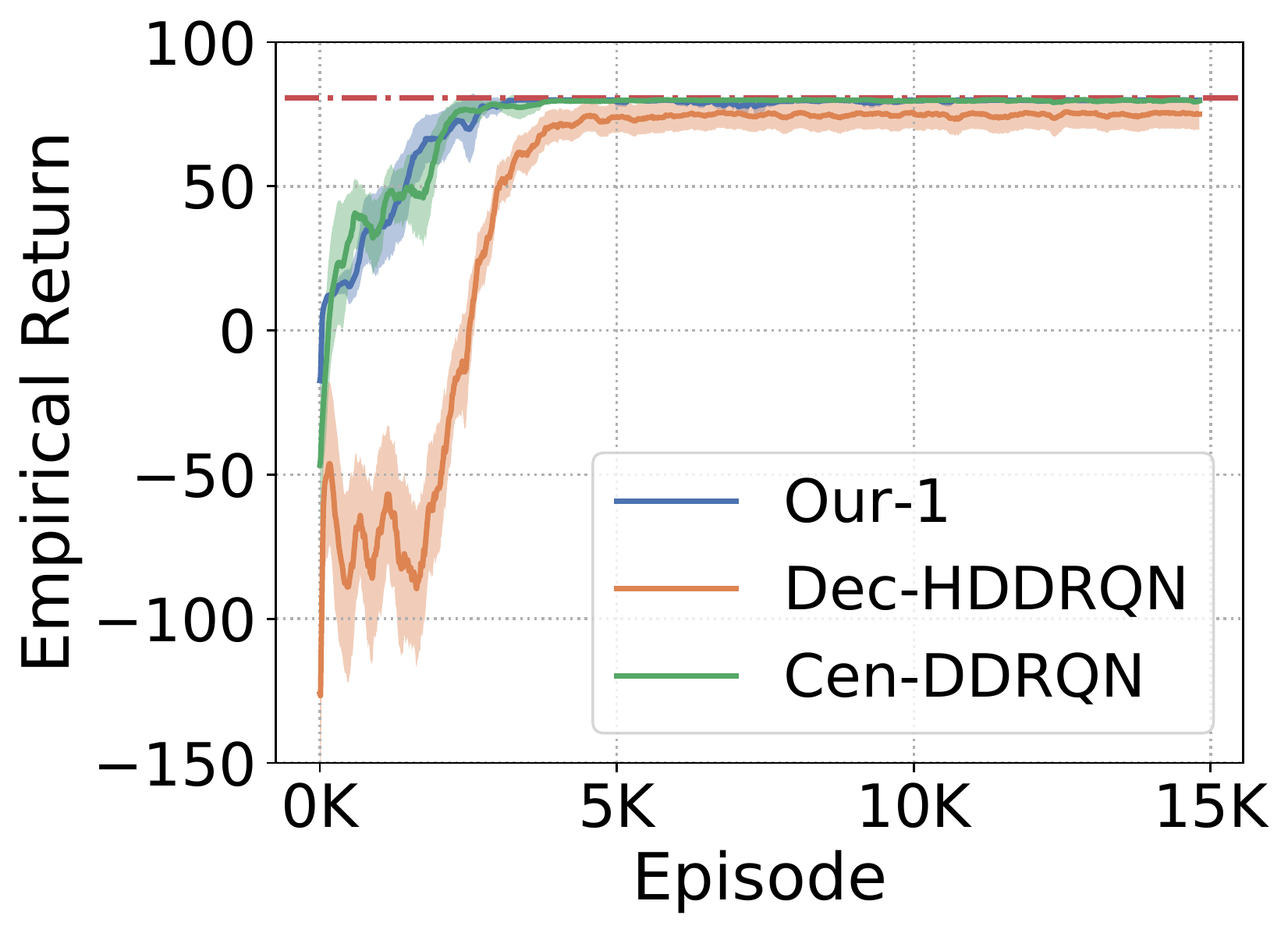}}
    ~
    \subcaptionbox{Grid world $30\times30$\label{bpma30}}
        [0.47\linewidth]{\includegraphics[height=3.1cm]{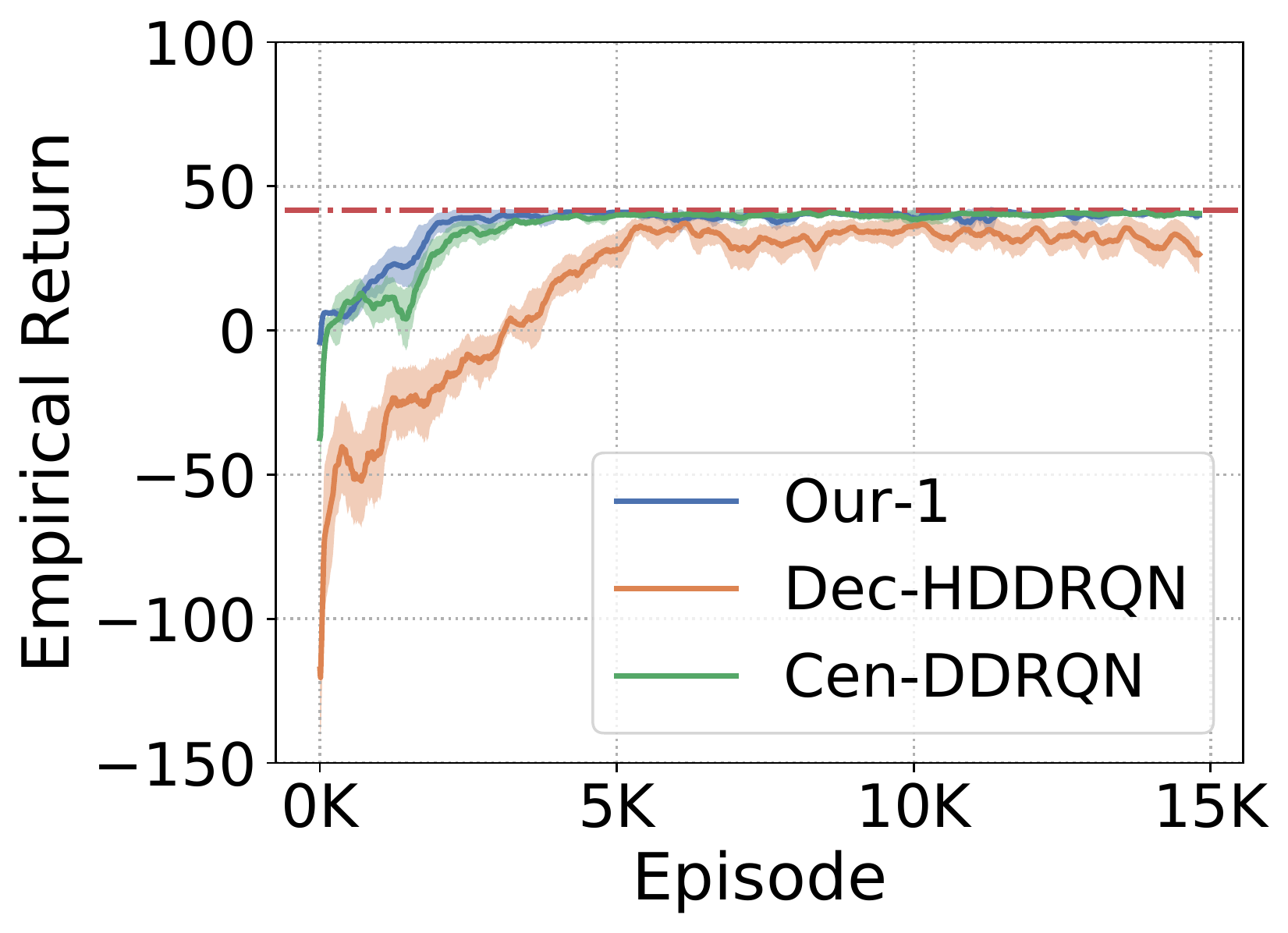}}
    \vspace{-1mm}
    \caption{Comparison of the average performance via three different learning approaches in BP domain.}
    \vspace{-5mm}
    \label{bpmaresults}
\end{figure}

We first evaluate our method \emph{MacDec-MADDRQN} (Our-1) with centralized $\epsilon$-greedy exploration in Box Pushing domain, and compare its performance with Dec-HDDRQN and Cen-DDRQN. In all three methods, the decentralized Q-net consists of two MLP layers, one LSTM layer~\cite{LSTM} and another two MLP layers, in which there are 32 neurons on each layer with Leaky-Relu as the activation function for MLP layers. The centralized Q-net has the same architecture but 64 neurons in the LSTM layer. The performance for two sizes of the domain is shown in Fig.~\ref{bpmaresults}, which is the mean of the episodic discounted returns ($\gamma=0.98$) over 40 runs with standard error and smoothed by 20 neighbors. The optimal returns are shown as red dash-dot lines. 

In both scenarios, the advantages of having the centralized $Q_\phi$ in the double-Q updating (Eq.~\ref{Condi}) is seen by it achieving similar performance to Cen-DDRQN and converging to the optimal returns earlier than Dec-HDDRQN. Furthermore, in the  bigger world space (Fig.~\ref{bpma30}), our method even leads to slightly faster learning than the fully centralized approach. This is because centralized Q-learning deals with the joint macro-observation and joint macro-action space, which is much bigger than the decentralized spaces from each agent's perspective. Our method has the key benefit of utilizing centralized information, but learning over a smaller space.   
 
 
\subsection{Results in the Warehouse Tool Delivery Domain}

We test our second proposed algorithm Parallel-MacDec-MADDRQN (Our-2) in this warehouse domain using the same evaluation procedure mentioned above.
\begin{wrapfigure}{R}{.27\textwidth}
    \centering
    \includegraphics[height=3.1cm]{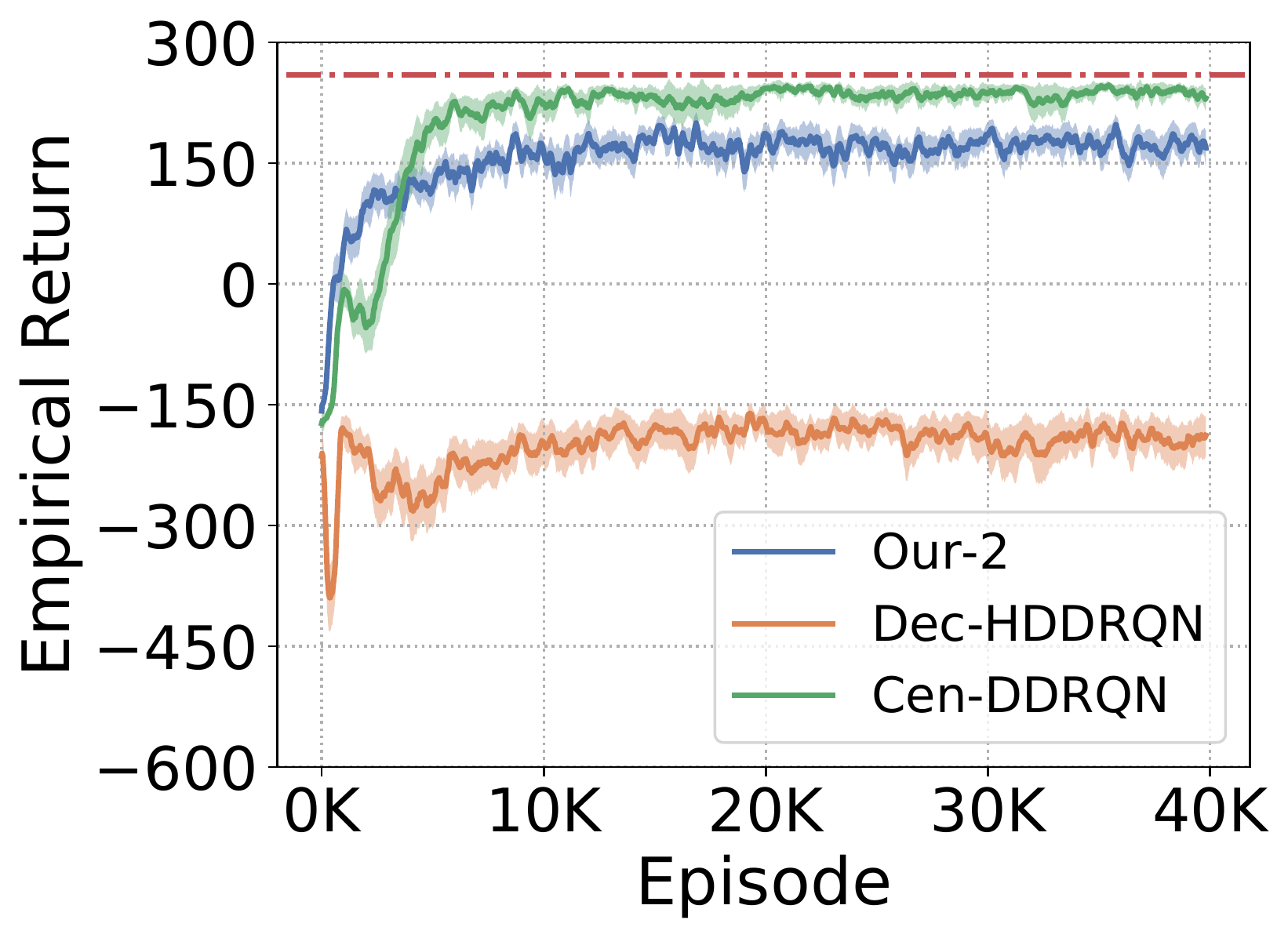}
    \vspace{-2mm}
    \caption{Performance of three different learning methods in WTD.}
    \vspace{-1mm}
    \label{result_wtd}
\end{wrapfigure}
The results shown in Fig.~\ref{result_wtd} are generated by using the same neural network architecture as the one adopted in the BP domain but with 32 neurons in each MLP layer and 64 neurons in LSTM layer for both the centralized Q-net and each decentralized Q-net because of the bigger macro-action and macro-observation spaces.  


The most challenging part in this domain is that robots need to reason about collaborations among teammates and which tool the human will need next. However, the gray robot, that plays the key role of finding the correct tool for delivery, does not have any knowledge about the human's need nor any direct observation of the human's status. Also, the mobile robots cannot observe each other. From the gray robot's perspective, the reward for its selection is very delayed, which depends on the mobile robots' choice and their moving speeds. For these reasons, each robot individually learning from local signals (in Dec-HDDRQN) leads to much lower performance but the centralized learner can achieve near-optimal results. 
Our approach achieves significant improvement while learning decentralized policies, but due to the limitation of local information, it inherently cannot perform as well as the centralized policy in such a complicated domain. Nevertheless, the near-optimal behaviors are still learned by our Parallel-MacDec-MADDRQN, which are presented in the real robot experiments (Section~\ref{HE}).       

\begin{wrapfigure}{L}{.25\textwidth}
    \centering
    \includegraphics[height=3.1cm]{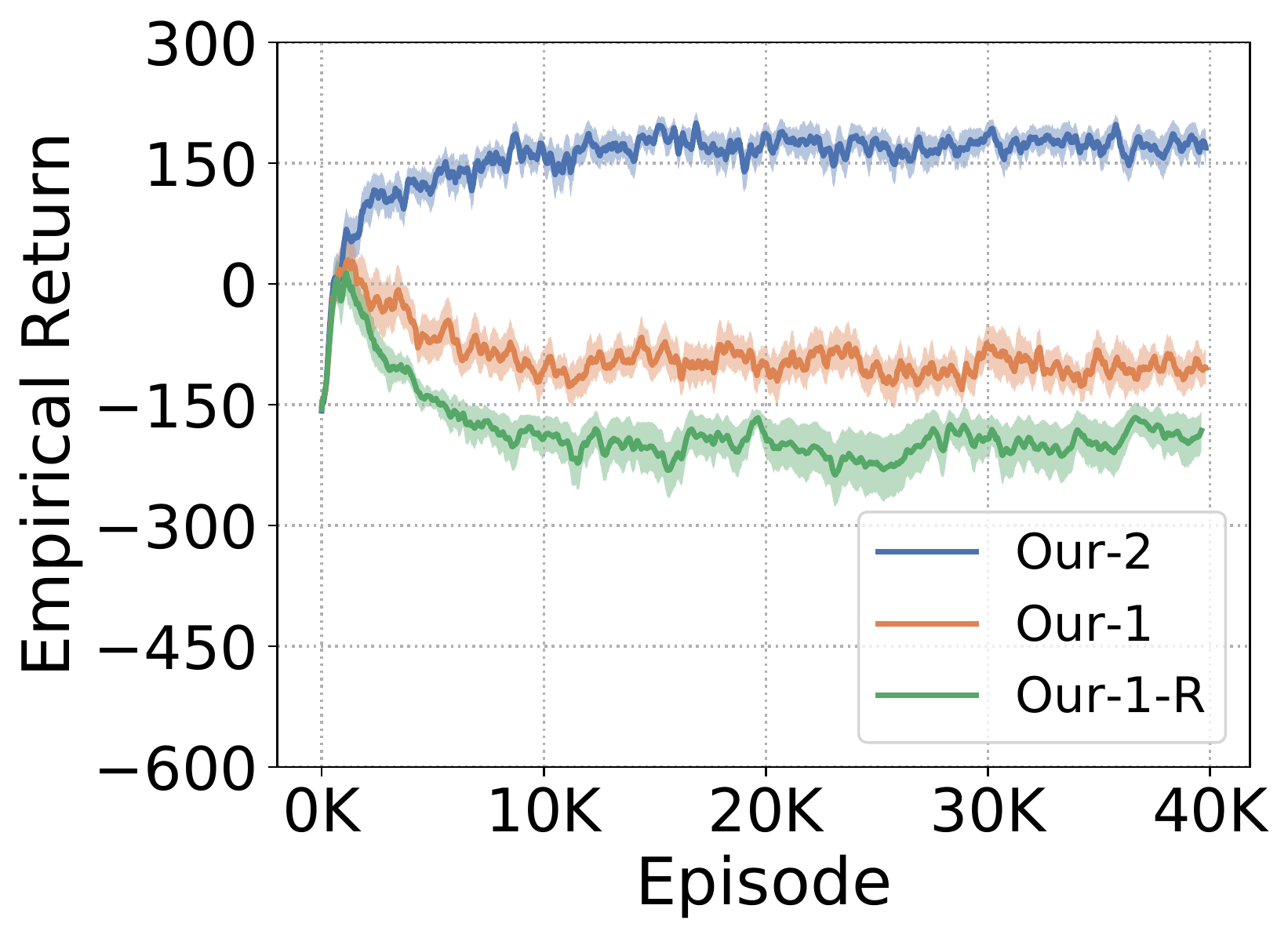}
    \vspace{-2mm}
    \caption{Results of ablation experiments in WTD.}
    \vspace{-2mm}
    \label{ablation}
\end{wrapfigure}

\vspace{-2mm}
We also conducted ablation experiments in WTD in order to investigate 1) the necessity of separately training the centralized Q-net and decentralized Q-nets in two environments by comparing Parallel-MacDec-MADDRQN (Our-2) with MacDec-MADDRQN with centralized exploration (Our-1); 2) the significance of including centralized $Q_\phi$ in double-Q updating to optimize each decentralized $Q_{\theta_i}$ (Eq.~\ref{Condi}) by performing Our-1 with regular deep double-Q learning (referred to Our-1-R). The results shown in Fig.~\ref{ablation} reveal that Our-2 outperforms other two ablations, which gives the affirmative answers to the above questions. 

\section{HARDWARE EXPERIMENTS}
\label{HE}

\begin{figure}[h!]
    \vspace{-2mm}
    \centering
    \includegraphics[height=3.2cm]{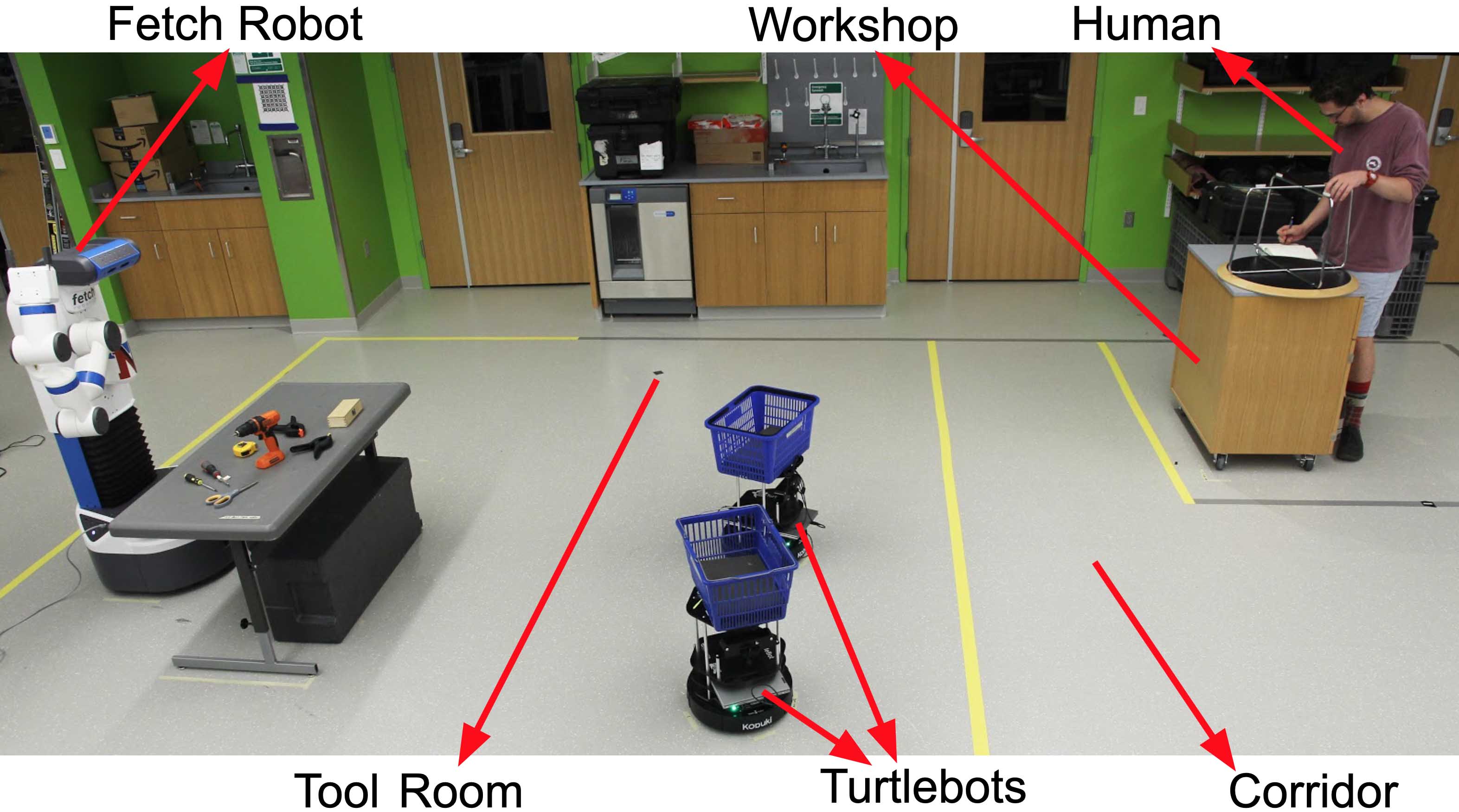}
    \vspace{-1mm}
    \caption{Hardware experiment setup.}
    \label{ExpStartFig}
    \vspace{-3mm}
\end{figure}

To verify that the learned decentralized policies in Parallel-MacDec-MADDRQN can effectively control a team of robots to achieve high-quality results in practice, we recreated the warehouse domain using three real robots: one Fetch robot~\cite{Wise:M} and two Turtlebots~\cite{Turtlebot} (Fig.~\ref{ExpStartFig}). 
A rectangle space with dimension \SI[]{5.0}{\meter} by \SI[]{7.0}{\meter} was taped to resemble the warehouse in the simulation (Section~\ref{Domains}). All the predefined waypoints and robots' initial positions were placed equal in ratio to the simulation. Also, the real-world human's task is to build a small table in the workshop, requiring three particular tools in the following order: a tape measure, a clamp and an electronic drill (from YCB object set~\cite{YCB}).  

Each robot had its own decentralized macro-observation space designed over ROS~\cite{ROS} services that kept broadcasting the signals about Turtlebots' locations, human's state (only accessible to the Turtlebot when it is located in workshop area), the status of each Turtlebot's basket, and the number of objects in the staging area (only observable in the tool room). Fetch's manipulation macro-actions are achieved by first projecting the point cloud data captured by Fetch's head camera into an OpenRAVE~\cite{Diankov:R} environment and performing motion planning using the OMPL~\cite{OMPL} library. The Turtlebot's movement macro-actions are controlled via the ROS navigation stack.   

\begin{figure}[t!]
    \centering
    \subcaptionbox{Fetch searches and stages the tape measure as T-1 approaches the table.\label{realrobots_a}}
        [0.45\linewidth]{\includegraphics[width=4.1cm, height=2cm]{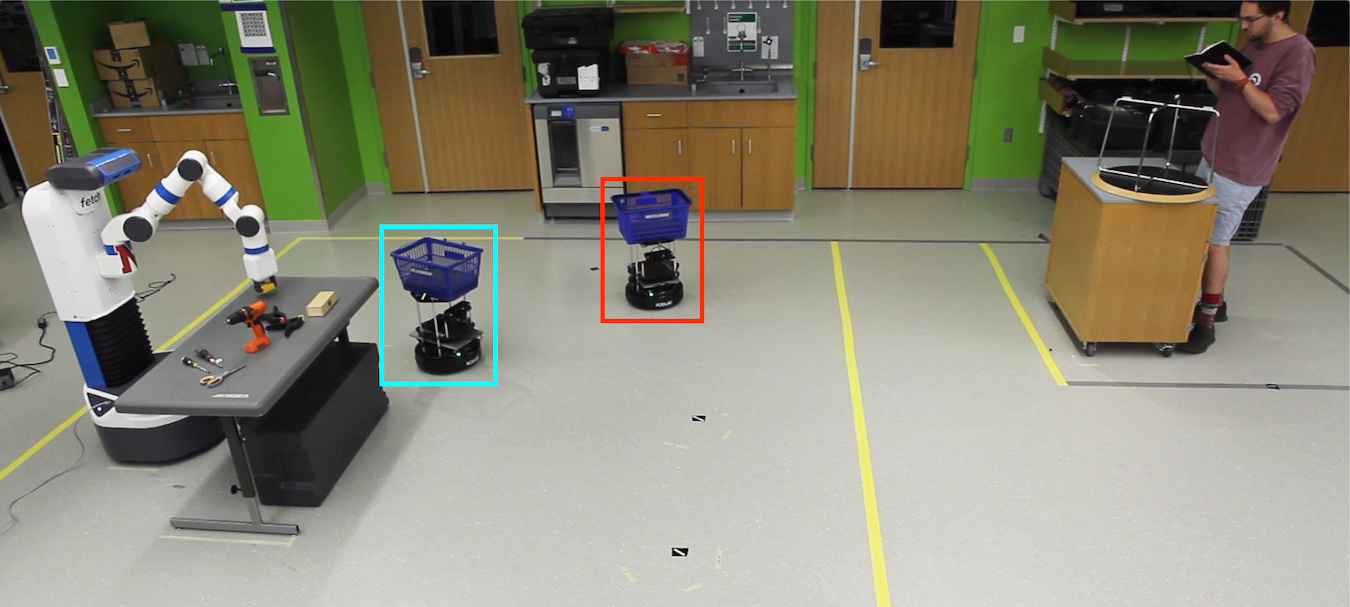}}
    ~
    \centering
    \subcaptionbox{Fetch sees T-1 arriving and passes it the tape measure, while T-0 reaches workshop and observes human's state.\label{realrobots_b}}
        [0.45\linewidth]{\includegraphics[width=4.1cm, height=2cm]{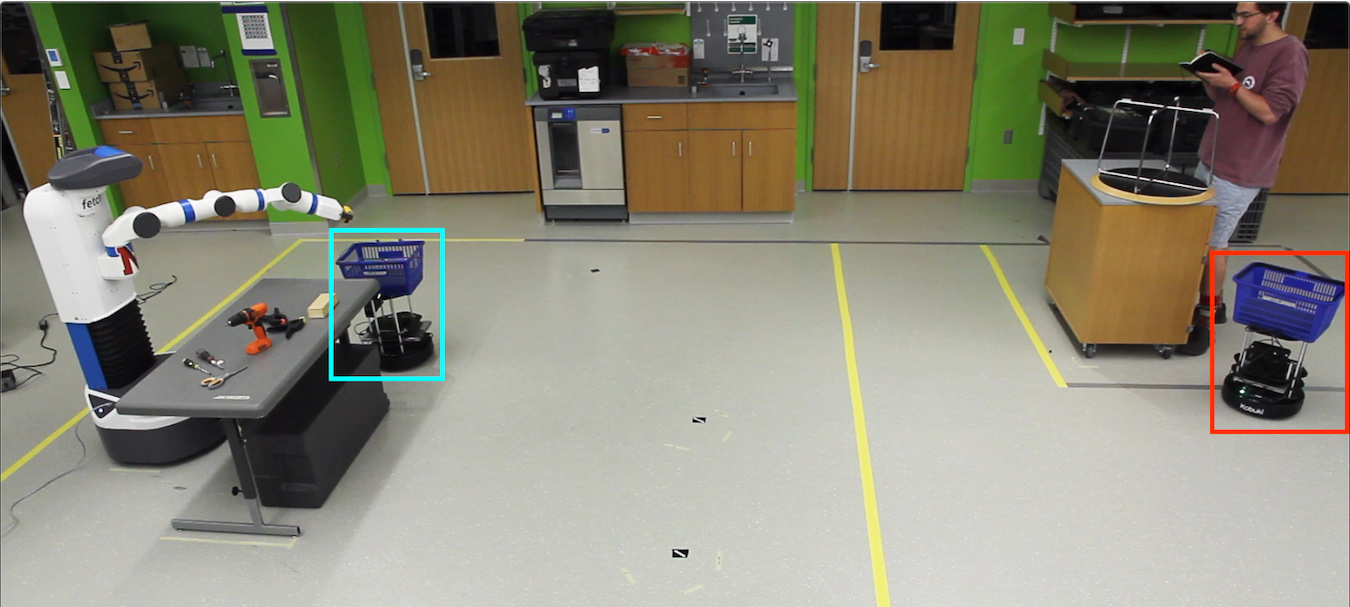}}
    ~
    \centering
    \subcaptionbox{T-1 observes the tape measure in its basket and moves to workshop, while T-0 goes back tool room and Fetch finds the clamp. \label{realrobots_c}}
        [0.45\linewidth]{\includegraphics[width=4.1cm, height=2cm]{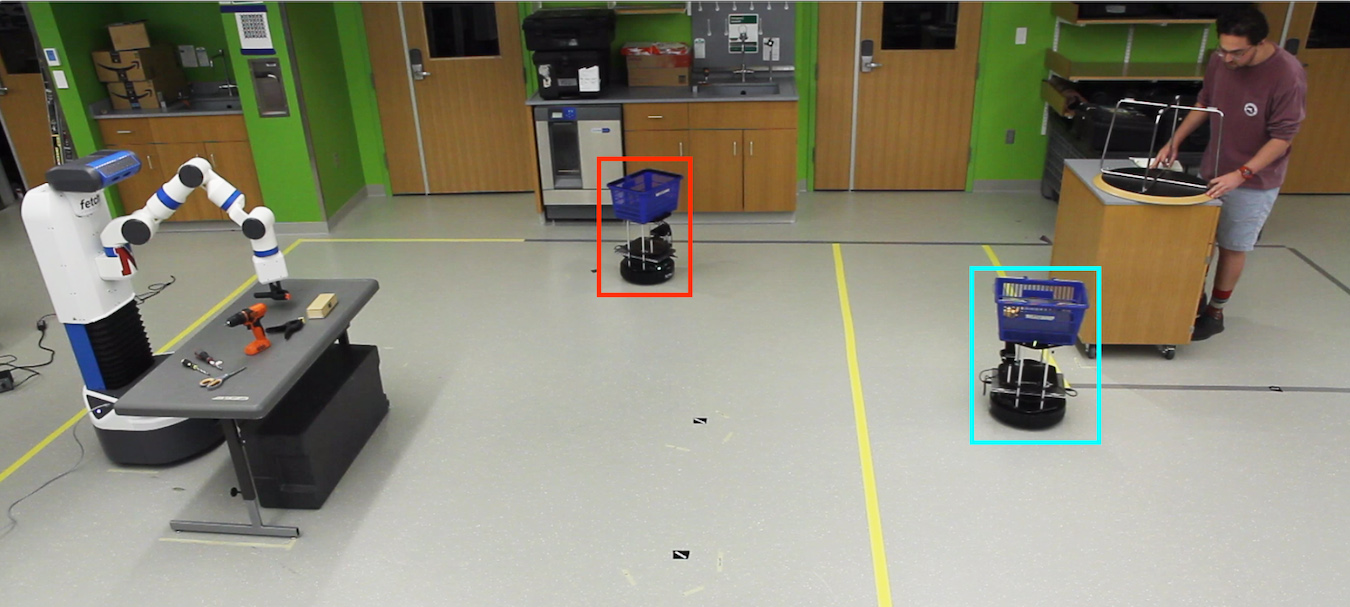}}
    ~
    \centering
    \subcaptionbox{T-1 deliveries the tape measure and T-0 runs to the table for the second tool, while Fetch notices no teammate around table yet.   \label{realrobots_d}}
        [0.45\linewidth]{\includegraphics[width=4.1cm, height=2cm]{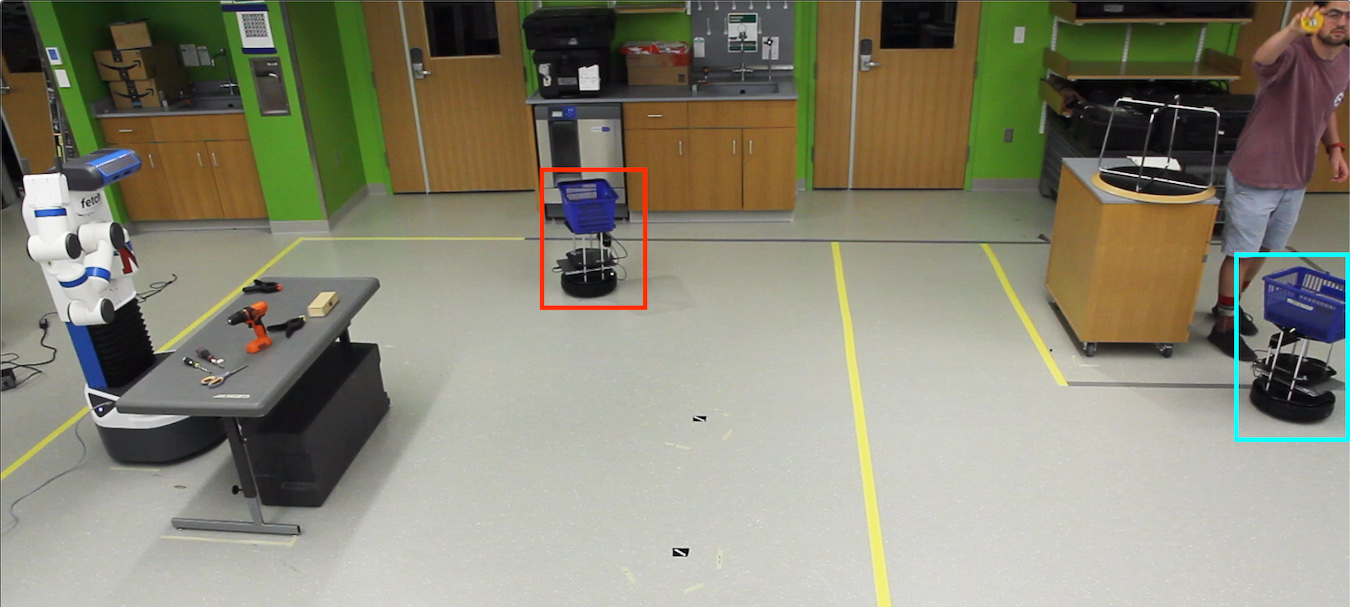}}
    ~
    \centering
    \subcaptionbox{Fetch grabs the electronic drill and stages it next to the clamp, while T-0 waits besides table and T-1 is coming back.\label{realrobots_e}}
        [0.45\linewidth]{\includegraphics[width=4.1cm, height=2cm]{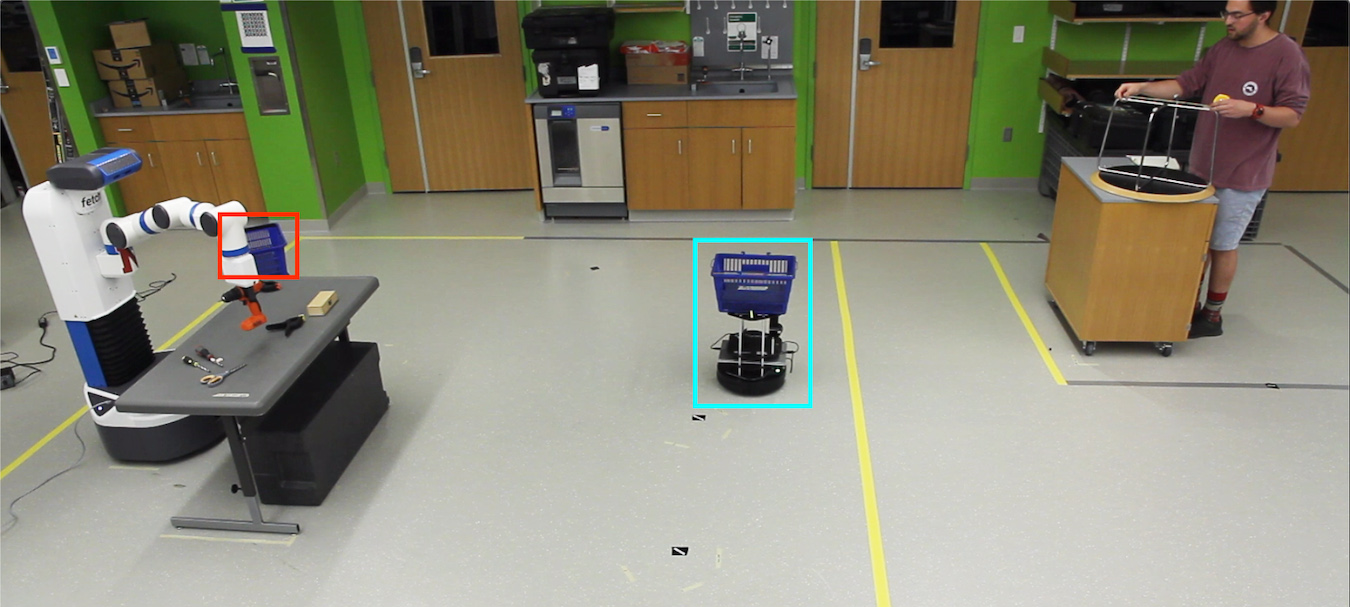}}
    ~
    \centering
    \subcaptionbox{Fetch observes T-0 has been ready there and passes clamp to it, in the mean time, T-1 arrives at the table.\label{realrobots_f}}
        [0.45\linewidth]{\includegraphics[width=4.1cm, height=2cm]{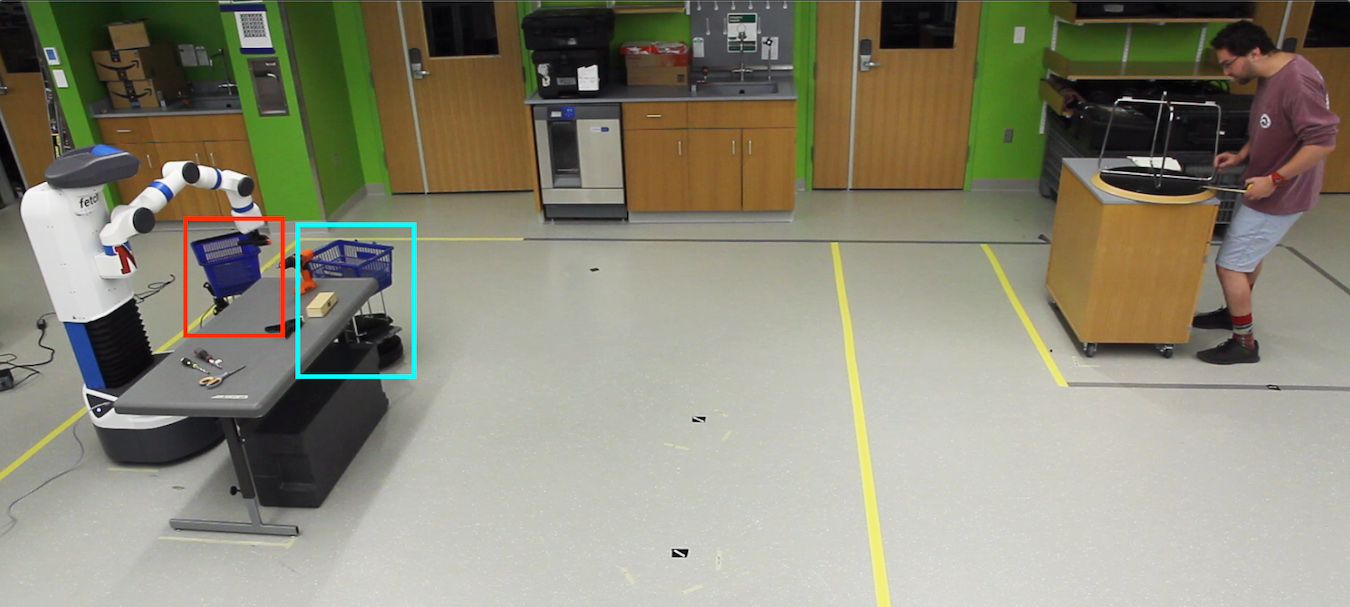}}
    ~
    \centering
    \subcaptionbox{T-0 immediately goes to send the 2nd tool and Fetch passes the last tool to T-1.\label{realrobots_g}}
        [0.45\linewidth]{\includegraphics[width=4.1cm, height=2cm]{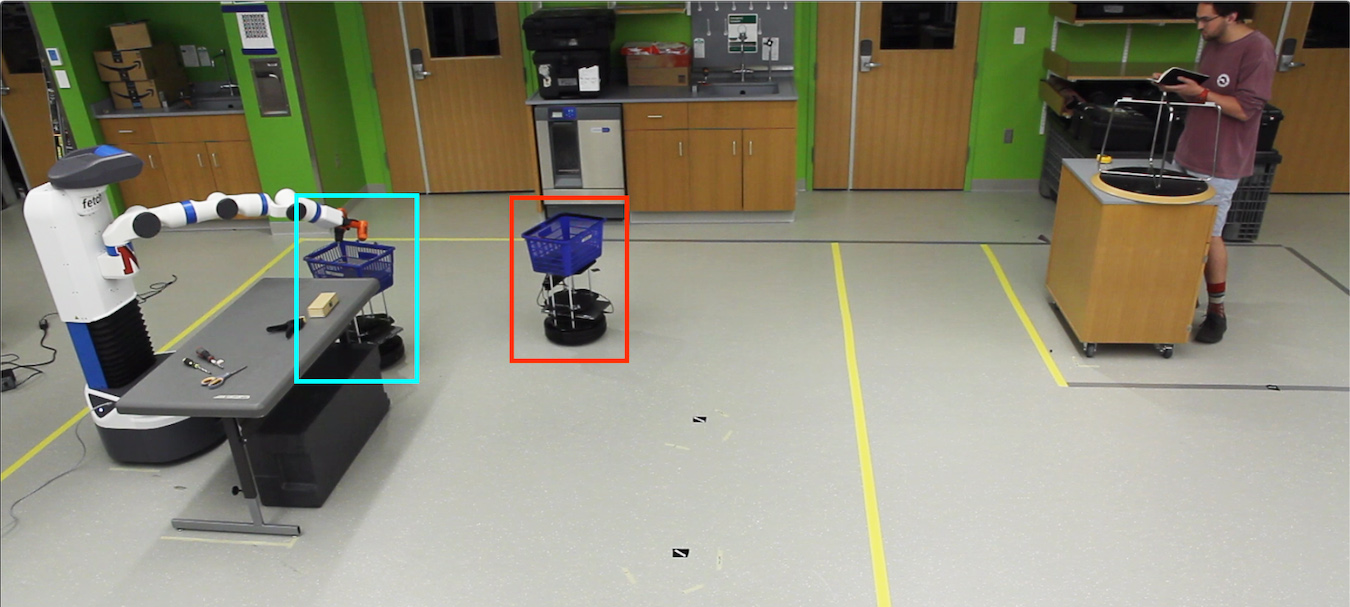}}
    ~
    \centering
    \subcaptionbox{Human gets the clamp from T-0, and T-1 is going to deliver the electronic drill.\label{realrobots_h}}
        [0.45\linewidth]{\includegraphics[width=4.1cm, height=2cm]{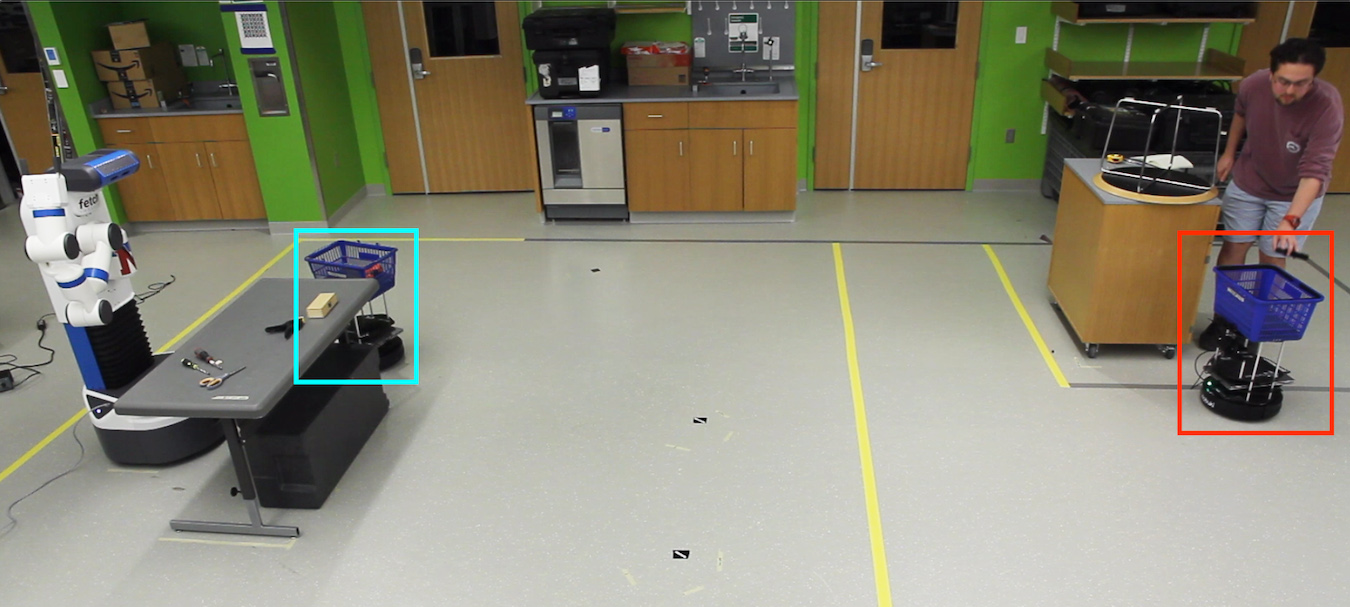}}
    ~
    \centering
    \subcaptionbox{The last tool is passed to the human by T-1 and the entire delivery task is completed.\label{realrobots_i}}
        [0.9\linewidth]{\includegraphics[width=4.1cm, height=2cm]{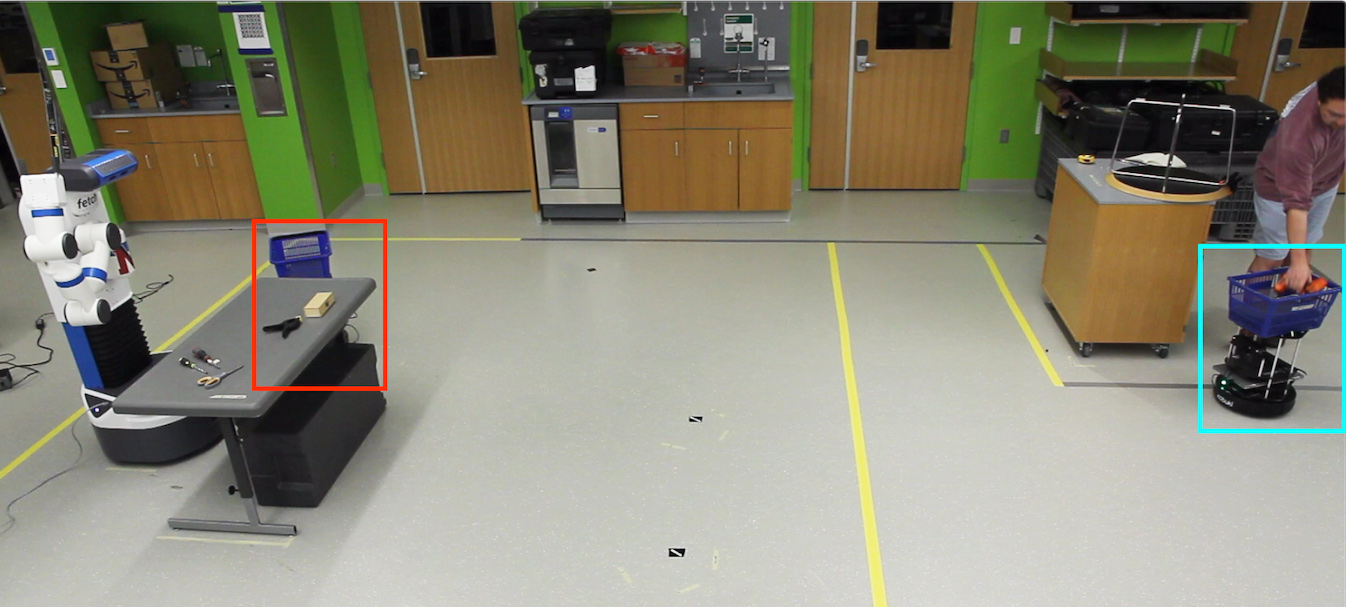}}

    \vspace{-1mm}
    \caption{Behaviors of robots running the decentralized policies (learned via Parallel-MacDec-MADDRQN) in the warehouse domain, where Turtlebot-0 (T-0) is bounded in red and Turtlebot-1 (T-1) is bounded in blue.} 
    \vspace{-5mm}
    \label{ExpStatesFig}
\end{figure}

Fig.~\ref{ExpStatesFig} shows the sequential cooperative behaviors performed by the robots. Although there is no direct interaction between the Fetch and the human, the trained policy learned the correct tools that the human needed and commanded the Fetch to find them in the proper order. Furthermore, the Fetch behaved intelligently such that: (a) Fig.~\ref{realrobots_c}-\ref{realrobots_e}, after placing the clamp into the staging area followed by observing no Turtlebot beside the table, it continued to look for the third object instead of waiting for Turtlebot-0 (bounded in red) to come over; (b) Fig.~\ref{realrobots_e}-\ref{realrobots_f}, after finding the electronic drill, it first passed the clamp (the correct second object that the human needed) to Turtlebot-0 who arrived the table ahead of Turtlebot-1(bounded in blue). Meanwhile, Turtlebots were also clever in such a way that: (a) they delivered the three tools in turn, instead of letting one of them deliver all the tools or perform delivery only after having all the tools in the basket which actually would make the human wait; (b) they directly went to the human for delivery after obtaining a tool from the Fetch without any redundant movement, e.g. going to the tool room waypoint again.

\section{CONCLUSION}

This paper introduces MacDec-MADDRQN and Parallel-MacDec-MADDRQN: two new macro-action-based multi-agent deep reinforcement learning methods with decentralized execution. These methods enable each agent's decentralized Q-net to be trained while capturing the effects of other agents' actions by using a centralized Q-net for decentralized policy updating. The results in the benchmark Box Pushing domain demonstrate the advantage of our methods where the decentralized training achieves equally good performance as the centralized one. Furthermore, the warehouse domain results confirm the benefits and the efficiency of our new double-Q updating rule. Importantly, a team of real robots running the decentralized policies learned via our method performed efficient and reasonable behaviors in the warehouse domain, which validates the usefulness of our macro-action-based deep RL frameworks in practice.   

{\bf \noindent Acknowledgements.}
This research was funded by ONR grant N00014-17-1-2072, NSF award 1734497 and an Amazon Research Award (ARA).

\newpage
\bibliographystyle{IEEEtran}
\bibliography{ref}

\end{document}